\documentclass[letterpaper, oneside, 10 pt, conference]{ieeeconf}
\IEEEoverridecommandlockouts
\usepackage[noadjust]{cite}
\usepackage{siunitx}
\usepackage{amsmath, bm, amssymb, mathrsfs, amsfonts}
\usepackage{algpseudocode}
\usepackage{algorithm}
\usepackage{graphicx}
\usepackage{textcomp}
\usepackage{xargs}
\usepackage{xcolor}
\usepackage{caption}
\usepackage{subcaption}

\makeatletter
\let\NAT@parse\undefined
\makeatother
\usepackage[hidelinks]{hyperref}

\newcommand{\correctionadj}[1]{#1}
\newcommand{\correctionnew}[1]{#1}
\long\def\invis#1{}   %make text invisible

\newtheorem{definition}{Definition}[section]

\newcommand{\pv}{\bm{\mathcal{P}}}
\newcommand{\pvd}{\pv'}
\newcommand{\pvdd}{\pvd'}
\newcommand{\qpos}{\bm{q}}
\newcommand{\qvel}{\dot{\qpos}}
\newcommand{\qacc}{\ddot{\qpos}}
\newcommand{\svel}{\dot{s}}
\newcommand{\sacc}{\ddot{s}}
\newcommand{\svelsq}{\svel^{2}}

\newcommand{\gctrl}{\bm{\tau}}
\newcommand{\ctrl}{\bm{u}}

\newcommand{\manipm}[1]{\bm{M}(#1)}

\newcommand{\manipc}[2]{\bm{C}(#1, #2)}
\newcommand{\manipg}[1]{\bm{G}(#1)}

\newcommand{\manipequlhs}{\manipm{\qpos} \qacc + \manipc{\qpos}{\qvel} \qvel + \manipg{\qpos}}

\newcommand{\tpas}{\bm{a}(s)}
\newcommand{\tpbs}{\bm{b}(s)}
\newcommand{\tpcs}{\bm{c}(s)}

\newcommand{\tree}{\mathcal{T}}

\DeclareMathOperator{\rank}{rank}
\DeclareMathOperator*{\argmin}{argmin}
\newcommand\norm[1]{\left\lVert#1\right\rVert}

\title{\LARGE \bf Path-Parameterised RRTs for Underactuated Systems}

\author{Damian Abood and Ian R. Manchester\thanks{{The authors are with the Australian Centre for Robotics and the School of Aerospace, Mechanical and Mechatronic Engineering, The University of Sydney, Australia. This research was supported in part by the ARC Research Hub for Intelligent Robotic Systems for Real-Time Asset Management (IH210100030). {\tt \small \correctionadj{\{damian.abood, ian.manchester\}@sydney.edu.au}}
}}}
\begin{document}
\maketitle
\thispagestyle{empty}
\pagestyle{empty}

% \textbf{Key:} \correctionnew{$\blacksquare$ - New Content},\,
% \correctionadj{$\blacksquare$ - Adjusted Content}

\begin{abstract}
We present a sample-based motion planning algorithm specialised to a class of underactuated systems using path parameterisation. The structure this class presents under a path parameterisation enables the trivial computation of dynamic feasibility along a path. Using this, a specialised state-based steering mechanism within an RRT motion planning algorithm is developed, enabling the generation of both geometric paths and their time parameterisations without introducing excessive computational overhead. We find with two systems that our algorithm computes feasible trajectories with higher rates of success and lower mean computation times compared to existing approaches.
% Success rates to go first.
% Faster
% Without introducing computational overhead / time-per-iteration
\end{abstract}

% \begin{IEEEkeywords}
% motion planning, path parameterisation, underactuated, random-search, dynamics
% \end{IEEEkeywords}
\section{Introduction}
% Importance of motion planning/underactuated systems
Generating dynamically feasible motions is fundamental to the planning and control of robotic systems~\cite{lavalle_planning_2006}. Underactuated systems such as flying and legged systems are fast becoming canonical to modern robotics, with growing applications towards remote inspection and surveillance of high-risk areas~\cite{somers_priority_2019, bellicoso_advances_2018}. A large challenge in motion planning for underactuated systems however is that their space of feasible motions is strongly constrained by their dynamics. Plans generated by motion planning algorithms must therefore exploit the natural dynamics of the system given this coupling~\cite{hereid_3d_2016}.
\subsection{Motion Planning Algorithms for Dynamic Systems}
% Local search methods
Local search strategies such as nonlinear programming involve the creation of mathematical programs from the associated kinodynamic constraints through transcription methods (e.g. direct collocation and multiple shooting~\cite{betts_practical_2010}). The generality of transcription methods in handling kinodynamic constraints has enabled such programs to produce collision-free optimal trajectories for a range of mechanical systems such as vehicles~\cite{tang_time-optimal_2019}, manipulators~\cite{howell_altro_2019} and legged systems~\cite{kuindersma_optimization-based_2016}. Generating plans within complex environments can however hinder these programs due to increased problem size and additional constraints, which lead to longer solve times and potentially an increase of local minima. 

% Sample-based approaches
Sample-based methods are designed with complex environments in mind, with algorithms such as Rapidly-Exploring Random Trees (RRTs)~\cite{kuffner_rrt-connect_2000} and Probabilistic Road Maps (PRMs)~\cite{simeon_manipulation_2004} demonstrating the ability to compute feasible solutions by exploring the available state space. Whilst feasible solutions are returned, they are almost always suboptimal unless additional routines are incorporated, as in RRT${}^{\star}$~\cite{karaman_sampling-based_2011}.

% Challenges from underactuated systems
\correctionadj{Continuous optimisation approaches naturally include underactuated systems due to their generic handling of dynamic constraints. These dynamic constraints can however be more challenging to satisfy in comparison to fully actuated systems, particularly if poor initialisations are made~\cite{kelly_introduction_2017}}. Whilst sampling-based methods also extend to underactuated systems, the difficulty of creating feasible trajectories between sampled states can lead to longer computation times to arrive at a feasible solution~\cite{lavalle_randomized_2001}. To ensure branches within the tree respect the underactuated dynamics, these algorithms must have a well-designed steering routine.

\subsection{Steering Methods}
% Control-based steering
Control-based steering methods drive a system to a target state through the design of an appropriate control law, \correctionadj{with the trajectory created by forward simulation} becoming a new branch in the tree. Control laws can be as trivial as sampling from a fixed set of inputs and durations~\cite{lavalle_randomized_2001}, encouraging exploration of the entire state space~\cite{pham_kinodynamic_2013}. Selecting from a fixed pool of inputs can however make steering exactly to a target challenging. Optimal control methods such as LQR~\cite{perez_lqr-rrt_2012} and nonlinear programming~\cite{stoneman_embedding_2014} can be used to provide this exact steering to a target state. These methods do incur a much larger computational demand, which can become increasingly expensive to solve with higher dimensions~\cite{pham_admissible_2017}.

% State-based steering
State-based steering can also be used in the case of fully actuated systems, where instead a path is created between two points first and is assessed for its feasibility afterwards via inverse dynamics~\cite{caron_completeness_2017}. This path-driven approach has enabled path parameterisation, a technique that decomposes a trajectory into its geometric and temporal components~\cite{bobrow_time-optimal_1985}, to be used for the dynamic evaluation of these created paths. Path parameterisation has been used extensively for creating time-optimal executions of fixed paths, with efficient algorithms catered to these problems using methods such as numerical integration~\cite{pham_general_2014}, reachability analysis~\cite{pham_new_2018} and continuous optimisation~\cite{verscheure_time-optimal_2009}. % Moving from time optimality to more general objectives requires the use of numerical optimisation, with 
The convexity of these fixed-path problems enables efficient implementations in conic programming~\cite{verscheure_time-optimal_2009} and interior-point methods~\cite{suleiman_time_2010}. Optimising the path in addition to its execution leads to a nonlinear program, which has been considered in~\cite{tang_time-optimal_2019} using a bilevel perspective. This method however requires paths to maintain feasibility at each iteration, which can be difficult for systems where their dynamics and geometric path are strongly coupled, such as those with underactuation.

% % Metric
% To explore the state space effectively, choosing appropriate vertices within the tree to extend from is essential. A distance metric is thus required to base this selection on, however even for low dimensional systems the design of a suitable metric can be non-trivial. It has been shown that for nonlinear system dynamics, Euclidean-based metrics often lead to poor performance~\cite{shkolnik_reachability-guided_2009}. Methods that employ local LQR policies can use their approximations for the system cost-to-go as a metric, which has shown to be an effective heuristic for steering to a point~\cite{perez_lqr-rrt_2012} given they reflect the reachability of the system. Euclidean-based distance metrics have however been demonstrated to work well under a state-based steering approach~\cite{pham_kinodynamic_2013} given they focus on the configuration of the system.

\subsection{Steering Methods for Underactuated Systems}
% Underactuated challenges for control-based steering
Compared to fully actuated cases, underactuated systems generally take longer computational time to compute solutions for when using control-based steering approaches, given most inputs for this class of system generate undesirable behaviours~\cite{lavalle_randomized_2001}. Optimal control methods can avoid this issue with their exact steering capabilities and have been shown to generate trajectories on low-dimensional models such as the acrobot~\cite{perez_lqr-rrt_2012}.

% Underactuated challenges for state-based steering
Underactuated systems are far more challenging for state-based steering, as the designed paths must satisfy their underactuated dynamics exactly to be feasible. Dynamically infeasible paths are not suitable for path parameterisation methods, as these approaches are designed for fully or redundantly actuated systems~\cite{pham_general_2014}. Although underactuated systems were not considered,~\cite{pham_kinodynamic_2013} presented an RRT-based algorithm \textit{AVP-RRT} that demonstrated success in finding feasible trajectories for a double pendulum system with severe torque limitations. Branches in the tree were accepted depending on their success of their path parameterisations by admissible velocity propagation (AVP). % Whilst underactuated systems can be handled under control-based steering, it is evident 

There are current limitations in adapting state-based steering approaches to underactuated systems, despite their prevalence in robotics. Path parameterisation has shown to be extendable to the class of underactuated systems with a single degree of underactuation, \correctionadj{where dynamically feasible trajectories are those that follow their ``decoupling vector fields''~\cite{bullo_kinematic_2001}}. The focus of~\cite{bullo_kinematic_2001} was however to generate paths that were feasible for all time parameterisations, limiting the choice of system to those without potential effects such as gravity. This class of underactuated system is well renowned for its application in gait generation in planar walking systems~\cite{grizzle_asymptotically_2001}, where ~\cite{manchester_real-time_2014} demonstrated an efficient and direct computation of the dynamic feasibility of generated paths through the path parameterisation of motion primitives. \correctionadj{Primitive-based planning does however restrict the complexity of output motions, with existing works requiring transitioning between primitives at zero-velocity states~\cite{bullo_kinematic_2001} or ensuring the primitives are simple enough to enable efficient quadrature computations~\cite{manchester_real-time_2014}. The same kinodynamic querying procedure of~\cite{manchester_real-time_2014} could be used to assess the feasibility of motions with more dynamic behaviours such as those created under sample-based planning, as highlighted in~\cite{pham_kinodynamic_2013}.}

\subsection{Proposed Contributions}
Under this motivation, we offer the following contributions within this paper
\begin{itemize}
\item We propose \textit{UA1-RRT}, a path parameterisation-based random-search algorithm specialised for the motion planning of \textit{underactuated degree-one} (UA1) systems.
\item  We present a state-based steering mechanism for the random sampling of geometric paths for UA1 systems, where no such method currently exists.
\item We provide a configuration-space distance metric for nearest neighbours selection, which allows the distinguishing of points in configuration space with different velocities.
\end{itemize}
\section{Preliminaries}
\subsection{Path Parameterising a Trajectory}
Consider an $n$-dimensional mechanical system with configuration $\qpos$ and $m$ inputs $\ctrl$ with dynamics provided by
\begin{equation}
\label{equ: manipulator equ}
\manipequlhs = \bm{B} \ctrl
\end{equation}
where $\bm{M} \in \mathbb{R}^{n \times n}$, $\bm{C} \in \mathbb{R}^{n \times n}$, $\bm{G} \in \mathbb{R}^n$, $\bm{B} \in \mathbb{R}^{n \times m}$. We assume that $\qpos$ contains only prismatic and revolute coordinates that can be readily mapped and ``unwrapped"~\cite{manchester_unifying_2018} respectively to $\mathbb{R}^n$. A trajectory $\qpos(t)$ can be path-parameterised such that an $n$-dimensional geometric path $\pv(s) : [0, 1] \rightarrow \mathbb{R}^n$ and corresponding monotonic time parameterisation  $s(t) \, : \, [0, T] \mapsto [0, 1]$ are created to achieve $\qpos(t) = \pv(s(t))$. Differentiating with respect to time, the generalised rates are consequently 
\begin{align}
    \label{equ: qvel_and_qacc}
    \qvel = \svel  \pvd ,\quad \qacc = \svelsq \pvdd + \sacc \pvd 
\end{align}
\correctionadj{Where $\pv' := \frac{d\pv}{ds}$ and $\dot{s} := \frac{ds}{dt}$ and likewise for other variables.}

With the linearity of $\bm{C}(\qpos, \, \qvel)$ in $\qvel$ for mechanical systems~\cite{pham_general_2014}, the substitution of \eqref{equ: qvel_and_qacc} into \eqref{equ: manipulator equ} provides a representation of the dynamics under $s(t)$ that can be factored into an affine expression in $\sacc$, $\svelsq$ and $\ctrl$
\begin{align}
    \label{equ: pp_manipulator_equ}
    \tpas \ddot{s} + \tpbs \dot{s}^2 + \tpcs = \bm{B} \ctrl
\end{align}
with coefficient vectors $\tpas,\,\tpbs,\,\tpcs \in \mathbb{R}^n$ given by
\begin{align*}
    \tpas &:= \bm{M}(\pv) \pvd\\
    \tpbs &:= \bm{M}(\pv)\pvdd + \bm{C}(\pv, \pvd)\pvd\\
    \tpcs &:= \bm{G}(\pv)
\end{align*}
where dependencies on $s$ and $t$ have been removed for brevity.

\correctionadj{We make the following definition regarding dynamic feasibility for systems under a time parameterisation $s(t)$:
\begin{definition}[Dynamic Feasibility]
\label{def: dynamic feasibility} 
A geometric path $\pv(s)$ is dynamically feasible for the system \eqref{equ: manipulator equ} if there exists a strictly monotonic time parameterisation $s(t) : [0, T] \mapsto [0, 1]$ with $s(0) = 0$ and $s(T) = 1$ such that \eqref{equ: pp_manipulator_equ} is satisfied for the duration of the trajectory, an equivalent condition being $\svel(t) > 0 \; \forall t \in (0, T)$. If no such $s(t)$ exists, the path is \textit{dynamically infeasible}.
\end{definition}}

\subsection{Underactuated Degree-One Systems}
Underactuated degree-one (UA1) systems are mechanical second-order systems where $\rank(\bm{B}) = n - 1$ in \eqref{equ: manipulator equ}. By an appropriate choice of coordinates, the dynamics of these systems can be written in the canonical form
\begin{equation}
\label{equ: manipulator equ ua1}
\manipequlhs = \begin{bmatrix}
0\\
\bm{\tau}
\end{bmatrix}
\end{equation}
where we now consider the generalised inputs $\gctrl \in \mathbb{R}^{n-1}$ to the system (i.e. $[0\; \gctrl^T]^T = \bm{B} \ctrl$) 

\section{Solving Path Parameterisations for UA1 Systems}
\label{sec: ud1}
\correctionadj{We consider the problem of generating trajectories  $\qpos(t) : [0, T] \rightarrow \mathbb{R}^n$ for UA1 systems within configuration space $\mathcal{Q} \subset \mathbb{R}^n$ connecting an initial state $(\qpos_0,\qvel_0)$ to a goal state $(\qpos_g, \qvel_g)$. These trajectories must satisfy the dynamics \eqref{equ: manipulator equ ua1} whilst respecting velocity and actuation bounds 
\begin{equation}
\label{equ: generalised bounds}
\begin{aligned}
\qvel_L \le \qvel(t) \le \qvel_U,\; \gctrl_L \le \gctrl(t) \le \gctrl_U
\end{aligned}
\end{equation}
The approach taken finds geometric path $\pv(s) : [0, 1] \rightarrow \mathbb{R}^n$ joining $\qpos_0$ to $\qpos_g$ within $\mathcal{Q}$ with corresponding time parameterisation $s(t) : [0, T] \rightarrow [0, 1]$ such that $\qpos(t) = \pv(s(t))$ satsifies \eqref{equ: manipulator equ ua1} and \eqref{equ: generalised bounds}.}

Under a path parameterisation $\qpos(t) = \pv(s(t))$, the dynamics \eqref{equ: manipulator equ ua1} can be split as follows
\begin{equation}
\label{equ: tp uad1 dynamics split}
\begin{bmatrix}
a_u\\
\bm{a}_a
\end{bmatrix} \sacc + 
\begin{bmatrix}
b_u\\
\bm{b}_a
\end{bmatrix} \svelsq + 
\begin{bmatrix}
c_u\\
\bm{c}_a
\end{bmatrix}= \begin{bmatrix}
0\\
\gctrl
\end{bmatrix}
\end{equation}
where the coefficient vectors $\bm{a},\bm{b},\bm{c}$ are partitioned into their actuated ($\bm{a}_a,\bm{b}_a,\bm{c}_a \in  \mathbb{R}^{n-1}$) and underactuated ($a_u, b_u, c_u \in \mathbb{R}$) components. The single passive degree of freedom in \eqref{equ: tp uad1 dynamics split} forces the evolution of $s(t)$ to be driven purely by the underactuated dynamics, that is $s(t)$ must satisfy 
\begin{equation}
\label{equ: uad1 dynamics}
a_u(s) \sacc + b_u(s) \svelsq + c_u(s) = 0
\end{equation}
Drawing parallels to virtual constraints~\cite{grizzle_asymptotically_2001} can this be viewed as the resulting zero-dynamics of the system when applying the virtual constraints $\qpos(t) = \pv(s(t))$. The constraint \eqref{equ: uad1 dynamics} imposed by underactuation implies $s(t)$ can be solved by numerical quadrature of \eqref{equ: uad1 first order}, only requiring a single forward pass as opposed to the forwards and backwards passes necessary for classical TOPP methods~\cite{bobrow_time-optimal_1985, pham_general_2014}.

Given an initial path velocity $\svel_0 \ge 0$, we choose to solve for the profile $\svelsq(s):= \theta(s)$ over the path domain, which when applying the identity $\sacc = \frac{1}{2}\frac{d}{ds} \svelsq$ presents a desirable form of \eqref{equ: uad1 dynamics} as a first order linear differential equation
\begin{equation}
\label{equ: uad1 first order}
a_u(s)\theta' +2 b_u(s) \theta +  2c_u(s) = 0
\end{equation}
where we can numerically integrate this system from $\theta_0 = \svelsq_0$ with a step size of $\Delta s$. At each point $s_i$ can the kinodynamic quantities $\qvel$, $\qacc$ and $\gctrl$ be recovered through \eqref{equ: qvel_and_qacc} and the remaining $n-1$ equations of \eqref{equ: tp uad1 dynamics split} using the computed $\theta(s_i)$, $\theta'(s_i)$ (using \eqref{equ: uad1 first order}) and $\pv(s_i)$.
\subsection{Dynamic Feasibility of a Path}
\label{sec: uad1 feasibility}
In UA1 systems, $\svel(t) < 0$ presents a physical significance given $\svel(t)$ is an indication of the available kinetic energy along the provided path~\cite{manchester_real-time_2014}. If $\svel(t) < 0$, then insufficient kinetic energy exists for the underactuated degree to complete motion along the path and instead ``falls back" on itself. Dynamic feasibility conditions under $\theta(s)$ are identical to $\svel(s)$ such that provided an initial path velocity $\svel(0) = \svel_0 \ge 0$, zero-crossings of $\theta(s)$ indicate the path is dynamically infeasible beyond the crossing. The forward integration scheme for $\theta$ enabled by \eqref{equ: uad1 dynamics} allows for zero-crossings in $\theta$ to be detected immediately, allowing rapid detection of path infeasibility and termination of the integration. As implied by \eqref{equ: uad1 first order}, zero-inertia points $s_z$ (i.e. $a_u(s_z) = 0$ in \eqref{equ: uad1 first order}) will always correspond to dynamic singularities where $\theta'(s_z)$ is undefined. To avoid unwanted behaviour in $\theta(s)$, we approximate $\theta'(s_z)$ at these points by finite-differencing local to $s = s_z$. 

\section{UA1-RRT Routine}
In this section, we present the main contribution of this paper, the UA1-RRT algorithm. We adopt a path-parameterised perspective to the trajectory planning problem, breaking the problem into the coupled subproblems of configuration space planning and time parameterisation of the resulting paths. 

\begin{figure}[t]
    \centering
    \includegraphics[width=\linewidth]{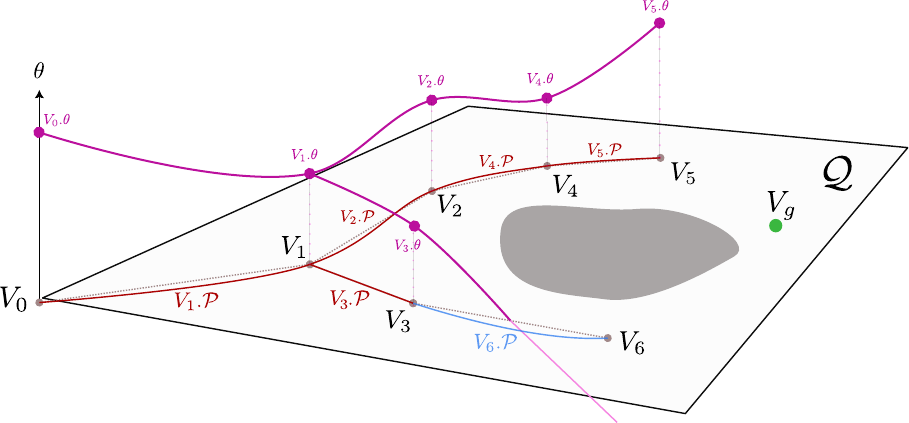}
    \caption{Visualisation of tree-growth in UA1-RRT,  \correctionnew{with edges $V_i.\pv$ connecting vertices $V_i$ in obstacle-free configuration space $\mathcal Q$. Path are added as edges to the tree if their path rate squared $\theta$ (magenta) remains positive (red) and discarded if a zero-crossing is encountered (blue).}}
    \label{fig: problem_representation}
\end{figure}

Our algorithm follows the general structure of AVP-RRT~\cite{pham_kinodynamic_2013}, using a standard RRT procedure~\cite{kuffner_rrt-connect_2000} to \correctionadj{create a collision-free geometric path in configuration space by iteratively constructing the path in segments and computing the trajectory along each segment via path parameterisation as per Sec. \ref{sec: ud1}}. \correctionnew{Whilst optimal tree-based searches exist, the typical rewiring step of optimal tree search methods~\cite{perez_lqr-rrt_2012} poses difficulty with underactuated systems due to the challenge in generating feasible motions between states.} The key difference of our proposed approach to AVP-RRT is that due to the constraint of underactuation in \eqref{equ: tp uad1 dynamics split}, the feasible velocity profile $\dot{s}(t)$ for a path segment $\pv(s)$ is unique as opposed to there being an interval of feasible path velocities~\cite{pham_admissible_2017}. UA1-RRT constructs configuration-space trees $\tree$ comprised of vertices $V$ that connect to other vertices with geometric paths (Fig. \ref{fig: problem_representation}), with trajectories generated through the path parameterisation method of Sec. \ref{sec: ud1}. A vertex $V$ therefore requires the following data: $V.\qpos$ - the location of the vertex in configuration space $\mathcal{Q}$; $V.\lambda$ - the parent vertex to $V$; $V.\pv$ - the connecting geometric path from $V.\lambda$ to $V$; $V.\theta$ - the path rate-squared at $V.\qpos$ when moving along $V.\pv$. 

\subsection{Algorithm Overview}
\correctionadj{We present our algorithm UA1-RRT in Algorithm \ref{alg: ua1-rrt}, with a graphical depiction of the process shown in Fig. \ref{fig: problem_representation}.}
The tree $\tree$ is initialised with a root vertex $V_0$ at the initial configuration $\qpos_0$. With desired initial velocity $\qvel_0$ and path velocity $\svel_0 \ge 0$, $V_0.\pvd(0)$ and rate-squared term $V_0.\theta$ are computed such that $\qvel_0 = \sqrt{V_0.\theta} V_0.\pvd(0)$. We arbitrarily select $V_0.\theta = \svelsq_0$ and $V_0.\pvd(0) = \qvel / \svel_0$ to satisfy this. In cases where $\qvel_0 = \bm{0}$, we also have the option of $\svel_0 = 0$ and $V_0.\pvd(0) \in \mathbb{R}^n$. \textproc{RandomConfiguration} (Alg. \ref{alg: ua1-rrt} L\ref{alg:line ua1-rrt random configuration}) generates a randomly sampled obstacle-free point $\qpos_r \in \mathcal{Q}$ for $\tree$ to extend towards. \textproc{ReachGoal} (Alg. \ref{alg: ua1-rrt} L\ref{alg:line ua1-rrt reach goal}) checks whether a vertex $V \in \tree$ is sufficiently close to the goal state, and if so, terminates the program. On termination, the final trajectory is computed by \textproc{ComputeTrajectory} (Alg. \ref{alg: ua1-rrt} L\ref{alg:line ua1-rrt compute trajectory}) which successively moves through each vertex of the branch connecting root to goal, concatenating the edges of each vertex to create the path of the trajectory. $\qpos(t)$ and $\qvel(t)$ are then computed by solving \eqref{equ: uad1 first order} for the provided path and initial path velocity. We now detail the \textproc{Extend} and \textproc{Steer} routines of UA1-RRT.

\begin{algorithm}[t]
    \caption{\textproc{UA1-RRT}}
    \label{alg: ua1-rrt}
    	\hspace*{\algorithmicindent} \textbf{Input:} $\qpos_0,\,\qvel_0 ,\, \svel_0 ,\, \qpos_g,\,\qvel_g$\\
    	\hspace*{\algorithmicindent} \textbf{Output:} Trajectory $\qpos(t), \qvel(t)$ joining $\qpos_0$ to $\qpos_g$
    \begin{algorithmic}[1]
    		\State $V_0.\qpos \gets \qpos_0,\,V_0.\theta \gets \svelsq_0,\,V_0.\pvd(0) \gets \qvel_0 / \svel_0$
        \State $\tree = \{V_0\}$
        \For{$i = 1, 2, \hdots, N$} \label{alg:line ua1-rrt max iterations}
        \State $\qpos_r = \textproc{RandomConfiguration}()$ \label{alg:line ua1-rrt random configuration}
        \State $V_e = \textproc{Extend}(\tree, \qpos_r)$ \correctionnew{\Comment Alg. \ref{alg: extend}} \label{alg:line ua1-rrt extend begin}
        \If{$V_e \ne \emptyset$}
        \State $\tree \gets \tree \cup \{V_e\}$
        \EndIf \label{alg:line ua1-rrt extend end}
        \If{$\textproc{ReachGoal}(\tree, \qpos_g, \qvel_g)$ successful} \label{alg:line ua1-rrt reach goal}
        \State $(\qpos(t), \qvel(t)) = \textproc{ComputeTrajectory}(\tree)$ \label{alg:line ua1-rrt compute trajectory}
		\State \Return{$(\qpos(t), \qvel(t))$}        		
		\EndIf
 
        \EndFor
    \end{algorithmic}
\end{algorithm}

\subsection{Extension}
\label{sec: extension}
The \textproc{Extend} procedure (Alg. \ref{alg: ua1-rrt} L\ref{alg:line ua1-rrt extend begin}) attempts to create a new vertex which is as close as possible to target configuration $\qpos_r$ to be added to $\tree$, with details of the routine shown in Alg. \ref{alg: extend}.
\begin{algorithm}[t]
    \caption{\textproc{Extend}}
    \label{alg: extend}
    \hspace*{\algorithmicindent} \textbf{Input:} Tree $\tree$, configuration $\qpos_r$ to extend $\tree$ towards\\
    	\hspace*{\algorithmicindent} \textbf{Output:} A new vertex to be added to $\tree$ or $\emptyset$
    	    \begin{algorithmic}[1]
		\Function{Extend}{$\tree, \qpos_r$}
        \State $X = \textproc{Nearest}_k(\mathcal{T}, \qpos_r), Y = \emptyset$ \label{alg: extend:line:nearest}
		\For{$x_i \in X$}
        \State $\delta \qpos = \qpos_r - x_i.\qpos$, $\hat{\delta \qpos} = \delta \qpos / \norm{\delta \qpos}$ \label{alg: extend distance limit begin}
		\If{$\norm{\delta \qpos}_2 > D_{\max}$} \label{alg:line: max steer}
		\State $\qpos_r \gets x_i.\qpos + D_{\max} \hat{\delta \qpos}$
		\EndIf \label{alg: extend distance limit end}
        \State $V_s = \textproc{Steer}(x_i, \qpos_r)$ \correctionnew{\Comment Alg. \ref{alg: steer}} \label{alg:line extend steer}
        \If{$V_s \ne \emptyset$ and $\textproc{IsObstacleFree}(V_s.\pv$)} \label{alg: extend obstacle free}
        \State $Y = Y \cup \{V_s\}$ \label{alg: extend y addition}
        \EndIf
        \EndFor
        \State \Return $\argmin_{y \in Y} \norm{\qpos_r- y.\pv(1)}_2$ \label{alg: extend return y}
        \EndFunction
    \end{algorithmic}
\end{algorithm}
To select suitable vertices within the tree to be extended from to the target point $\qpos_r$, \textproc{Nearest}${}_k$ (Alg. \ref{alg: extend} L\ref{alg: extend:line:nearest}) returns a set of $k$ vertices $X \subset \tree$ which are closest to $\qpos_r$ by a distance metric. The distance metrics in AVP-RRT used the Euclidean distance within the configuration space, with the option to include the final path orientation~\cite{pham_kinodynamic_2013}. 

Where repetitive motions or high-velocity manoeuvres are required, incorporating the orientation/velocity in these metrics is essential, given vertices that are close together in configuration space but with different velocities are indistinguishable. We propose the metric $d_p$ that considers the difference between $\qpos_r$ and the projection of a vertex's configuration $V_i.\qpos,\, V_i \in \tree$ under its current velocity for a user-defined duration $\gamma \in \mathbb{R}^+$. We define this projected point $\qpos_p \in \mathbb{R}^n$ as
\begin{equation}
\label{equ: projected point}
\qpos_p = V_i.\qpos + \gamma \sqrt{V_i.\theta}V_i.\pv'(0)\\
\end{equation}
which allows us to write the metric $d_p$ (normalised) as
\begin{equation}
\label{equ: projected metric}
d_p = \sum_{i = 1}^n \frac{\norm{q_{i,r} - q_{i,p}}^2}{\norm{q_{i,\max}}^2}
\end{equation}
This metric is well suited towards state-based steering as it will prioritise vertices in $\tree$ that have the most potential for the system to reach $\qpos_r$ from (or at least move in the direction of $\qpos_r$) based on their dead reckoning over horizon $\gamma$ (i.e. $\qpos_p$).

Each vertex $x_i \in X$ is then steered towards the target point $\qpos_r$. A maximum extension distance $D_{\max}$ is enforced to regulate tree growth such that extensions for the tree do not exceed this distance (Alg. \ref{alg: extend} L\ref{alg: extend distance limit begin}-\ref{alg: extend distance limit end}). \textproc{Steer} (Alg. \ref{alg: extend} L\ref{alg:line extend steer}) returns a vertex $V_s$ with a feasible edge connected to $x_i \in \tree$ extending as close as possible to $\qpos_r$ (details in Sec. \ref{sec: method steering}). 

\textproc{IsObstacleFree} (Alg. \ref{alg: extend} L\ref{alg: extend obstacle free}) performs collision checking of the path $V_s.\pv$ with any configuration-space obstacles \correctionnew{(shaded region of $\mathcal{Q}$ in Fig. \ref{fig: problem_representation})}. If no collisions are made, we add $V_s$ to the candidate vertex set $Y$ (Alg. \ref{alg: extend} L\ref{alg: extend y addition}). After all vertices in $X$ are steered towards, the closest vertex in $Y$ to $\qpos_r$ by Euclidean distance is returned as the new vertex to be added to $\tree$ (Alg. \ref{alg: extend} L\ref{alg: extend return y}).

To encourage growth towards the goal region, we periodically call \textproc{Extend} (Alg. \ref{alg: ua1-rrt} L\ref{alg:line ua1-rrt extend begin}-\ref{alg:line ua1-rrt extend end}) with $\qpos_r = \qpos_g$~\cite{pham_kinodynamic_2013}. As the tree grows closer to the goal region over the duration of the program, the projected point $\qpos_p$ in \eqref{equ: projected metric} will likely overshoot the goal when using a fixed $\gamma$. To account for this, we set $\gamma = 0$ for $d_p$ in \eqref{equ: projected metric} for this extension process to base distance purely on proximity to the goal in $\mathcal{Q}$. Together with the original \textproc{Extend} procedure, we offer a search method that encourages dynamically conscious tree growth in configuration space coupled with refined goal-reaching capabilities once significantly extended. 
\subsection{Steering}
\label{sec: method steering}
Our state-base steering approach motivated by the use of the path parameterisation allows us to create connecting paths $\pv(s)$ first and then determine whether they can be added to $\tree$ if their corresponding rate-squared profile $\theta(s)$ (by \ref{sec: ud1}) complies with kinodynamic constraints \eqref{equ: manipulator equ} and \eqref{equ: generalised bounds}. Using a path parameterisation also has the additional freedom that the execution time of a path is encoded by the profile $\theta(s)$, normally a difficult parameter to tune in the RRT method~\cite{lavalle_randomized_2001} without resorting to a sophisticated control policy.

To create geometric paths $\pv(s)$, we require they be smooth with $\mathcal{C}^1$ continuity such that $\qpos(t)$ and $\qvel(t)$ are continuous. Adding a new vertex to the tree must therefore ensure its path created by \textproc{Steer} (Alg. \ref{alg: extend} L\ref{alg:line extend steer}, Alg. \ref{alg: steer}) is $\mathcal{C}^1$-continuous to the path of its parent vertex. An exception to these continuity requirements is where $\qvel \approx 0$ at an existing vertex in the tree. In these cases, the initial tangent vector of the new vertex's path can ignore this continuity requirement, provided we have $\theta = 0$ for this point.

\textproc{GeneratePath} (Alg. \ref{alg: steer} L\ref{alg:line steer generate path}) creates paths $\pv(s)$ which meet these continuity requirements. Paths are created that connect a vertex $x_0 \in \tree$ to configuration $\qpos_r$ such that $\pv(s)$ is tangent to the tree at $x_0.\qpos$. Whilst any polynomial of degree three or higher meets this criterion, \textproc{GeneratePath} selects paths from the family of cubic polynomials with fixed end-point positions and initial gradients. These polynomials have one remaining degree of freedom which we use to parameterise this family of paths, which we choose to be the gradient at the end of the path. Randomly selecting the gradient within a user-defined range is performed on each call to \textproc{GeneratePath}, providing a randomised $\mathcal{C}^1$-continuous path $\pv(s)$ steering from $x_0.\qpos$ to $\qpos_r$. 

\begin{algorithm}[t]
    \caption{\textproc{Steer}}
    \label{alg: steer}
    \hspace*{\algorithmicindent} \textbf{Input:} Vertex $x_0 \in \tree$, target configuration $\qpos_r$\\
    	\hspace*{\algorithmicindent} \textbf{Output:} A vertex at $\qpos_r$ with parent $x_0$ or $\emptyset$
    	    \begin{algorithmic}[1]
		\Function{Steer}{$x_0,\,\qpos_r$}
        \State $Y = \emptyset$
        \For{$i = 1,2, \hdots, N_{rndm}$}
        \State $\pv(s) = \textproc{GeneratePath}(x_0, \qpos_r)$ \label{alg:line steer generate path}
        \State $(s^\star, \theta^\star) = \textproc{PathProfile}(\pv(s), x_0.\theta)$ \label{alg:line steer path profile} \Comment{\correctionnew{Sec. \ref{sec: ud1}}}
		\If{$s^\star \ge s^\dag$}
        \State $y.\pv \gets \{\pv(s) \; | \; \forall s \in [0, s^*]\}$
        \State $y.\theta = \theta^\star$
        \State $Y = Y \cup \{y\}$
        \EndIf
        \EndFor
        \State \Return{$\argmin_{y \in Y} \norm{\qpos_r- y.\pv(1)}_2$}
        \EndFunction
    \end{algorithmic}
\end{algorithm}

As shown in \ref{sec: ud1}, the velocity profile $\qvel(t)$ for UA1 systems is determined purely by the path $\pv(s)$ and initial velocity $\svel_0$. Ensuring bounds such as \eqref{equ: generalised bounds} are respected therefore requires the appropriate $\pv(s)$ if $\svel_0$ is already given. From our proposed \textproc{GeneratePath}, it is expected that most paths will only partially respect these bounds \eqref{equ: generalised bounds}. Despite this limitation, we can exploit the uniqueness of $s(t)$ for a given path by truncating paths $\pv(s)$ beyond a point $s^\star > 0$ (i.e. $\pv(s) \gets \{\pv(s) \; | \; \forall s \in [0, s^\star] \}$). \correctionadj{We choose the point $s^\star$ such that the truncated path is dynamically feasible by \eqref{equ: uad1 dynamics} whilst adhering to the imposed kinodynamic bounds \eqref{equ: generalised bounds}. Formally, we define the set of admissible path rates that respect \eqref{equ: generalised bounds} as $\mathcal{S}_{\pv}(s)$, given by
\begin{equation}
\mathcal{S}_{\pv}(s) = \{(\svel, \sacc) \; | \; \qvel_L \le \qvel(s) \le \qvel_U, 
\gctrl_L \le \gctrl(s) \le \gctrl_U \}
\end{equation}
where $\qvel(s)$ and $\gctrl(s)$ are computed with \eqref{equ: qvel_and_qacc} and \eqref{equ: tp uad1 dynamics split} respectively. With this, we can define $s^\star$ as the largest value in $s \in [0, 1]$ such that $\pv(s)$ over the interval $[0, s^\star]$ is dynamically feasible by \eqref{equ: uad1 dynamics} and the rates of its corresponding time parameterisation $s(t)$ are contained within $S_{\pv}(s)$.} Our definition for $s^\star$ does not include paths that encounter dynamic infeasibility (Def. \ref{def: dynamic feasibility}), as we believe creating paths that terminate at zero crossings of $\theta$ lead to poor future extensions since they indicate paths of insufficient kinetic energy to complete future motion~\cite{manchester_real-time_2014}. \correctionnew{This is demonstrated in Fig. \ref{fig: problem_representation}, where the branch from $V_3$ to $V_6$ is discarded given a zero crossing of $\theta$ is encountered along $V_6.\pv$}. The search for $s^\star$ is performed within the numerical integration of $\theta(s)$ (\textproc{PathProfile}) and is returned upon termination of the integration.

A cut-off threshold $0 < s^\dagger \le 1$ is set so that paths with $s^\star < s^\dagger$ are discarded to avoid slow tree growth. In our implementation, we choose the fairly conservative value $s^\dagger = 5 \Delta s$. To increase the likelihood of returning a vertex with a feasible path, we create $N_{rndm}$ random paths to assess. After all $N_{rndm}$ paths are computed and parameterised, \textproc{Steer} returns the vertex whose path terminates closest to $\qpos_r$ in the Euclidean sense. 

\subsection{Terminal Conditions}
\label{sec: terminal conditions}
$\textproc{ReachGoal}$ (Alg. \ref{alg: ua1-rrt} L\ref{alg:line ua1-rrt reach goal}) determines if any vertex in $\tree$ is within a user-defined distance of the goal $(\qpos_g,\,\qvel_g)$ and signals termination of the program if so. Given the terminal state $\qpos = V.\qpos$ and $\qvel = \sqrt{V.\theta} V.\pvd(1)$ of a vertex $V \in \tree$, normalised goal metrics for prismatic ($T$) and revolute ($R$) coordinates are computed as 
\begin{equation}
d_{g,i} = \begin{cases}
\frac{\norm{q_{g,i} - q_i}}{\norm{q_{i,\max}}} &\quad i \in T\\ 
(1 - \cos(q_{g,i} - q_i)) &\quad i \in R
\end{cases}
\end{equation}
Together with the normalised velocity error, the overall distance-to-goal is then
\begin{equation}
\label{equ: distance to goal}
d_{goal} = \frac{1}{2n} \bigg(\sqrt{\sum_{i=1}^{n} d^2_{g,i}} + \sqrt{\sum_{i=1}^{n} \frac{(\dot{q}_{g,i} - \dot{q}_i)^2}{\dot{q}_{i, \max}^2}}\bigg)
\end{equation}
For a distance threshold $\epsilon_g$, the tree has reached the goal if a vertex $V \in \tree$ returns a goal metric $d_{goal} \le \epsilon_g$. We select a value of $\epsilon_g = 10^{-2}$ to provide sufficient accuracy in the resulting trajectories akin to AVP-RRT~\cite{pham_kinodynamic_2013}. 

\section{Simulation Experiments}
\begin{figure}[t]
    \centering
	\begin{subfigure}[b]{0.35\textwidth}
         \centering
         \includegraphics[width=0.6\linewidth]{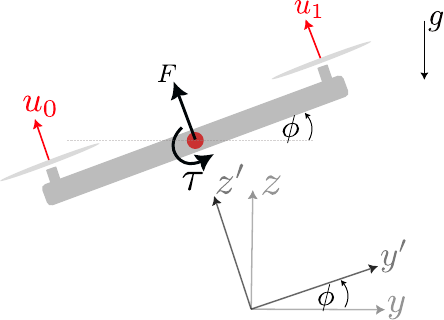}
         \caption{UAV model.}
         \label{fig: uav diagram}
     \end{subfigure}    
     \begin{subfigure}[b]{0.35\textwidth}
         \centering
         \includegraphics[width=0.6\linewidth]{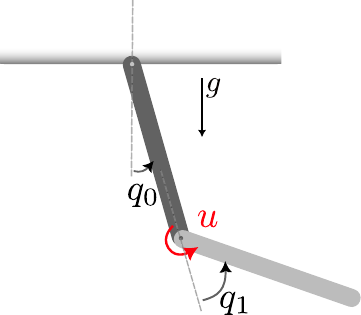}
         \caption{Acrobot model.}
         \label{fig: acrobot diagram}
     \end{subfigure} 
    \caption{\correctionadj{Underactuated degree-one (UA1) example systems.}}
    \label{fig: diagrams}
\end{figure}

We considered two UA1 systems for the demonstration of our method (Fig. \ref{fig: diagrams}), a planar UAV that must navigate a $4m$ obstacle-filled tunnel and an acrobot performing a swing-up procedure. 

\textit{UAV: }We chose the physical parameters $m = 0.1\,kg$, $I_{xx} = \num{1e-4}\,kgm^2$, with uni-directional thruster limits of $1\,N$. We imposed maximum translational and rotational velocities of $20\,ms^{-1}$ and $50\,rad/s$ respectively. The dynamics of the UAV within the inertial frame ($y,z, \phi$) along with inputs $(u_0, u_1)$ (Fig. \ref{fig: uav diagram}) does not readily produce dynamics of the form \eqref{equ: tp uad1 dynamics split}. Instead, we consider the dynamics in the body frame ($y',z', \phi$) with the net translational and rotational forces on the body $(\bm{F}, \bm{\tau})$ as inputs to the system. From this, the passive degree of freedom is the lateral axis $y'$ as no input can create additional acceleration in this direction. We specify the endpoints of the trajectory to be the start and end of the tunnel with target velocities $\qvel_0 = \qvel_g = \bm{0}$. For tree search parameters, we selected $D_{\max} = 1.0$ and $\gamma = 1.0$.

\textit{Acrobot: }The physical parameters for the acrobot are based on those of \cite{pham_kinodynamic_2013} now with a purely passive shoulder and an actuated elbow with torque limit $|u| \le 50\,Nm$. We have initial conditions $\qpos_0 = \qvel_0 = \bm{0}$ and velocity bounds $|\qvel_{0, 1}| \le 50\,rad/s$. Our acrobot can swing any number of rotations around its shoulder joint but with constraints on the elbow joint to reside in $q_2 \in [-\pi, \pi]$. Using the unwrapped space~\cite{manchester_unifying_2018} for $\qpos$, we choose a finite number of goal states to reach, $\qpos_g = [\pi + 2 \pi p, 0]^T$, $\qvel_g = \bm{0}$ with $p = 0, \pm 1, -2$. For tree search parameters, we selected $D_{\max} = 2.0$ and $\gamma = 0.1$.

We implemented the UA1-RRT routine within C++. We compare our approach to two other methods, also implemented in C++. The first is an adaptation of the AVP-RRT algorithm~\cite{pham_kinodynamic_2013} performing the same procedure in Algorithm \ref{alg: ua1-rrt} expect for the path profile step (Alg. \ref{alg: steer} L\ref{alg:line steer path profile}) being the AVP algorithm provided by the TOPP library~\cite{pham_general_2014}. The second method is a standard RRT routine with $k$ nearest neighbours (\textit{KNN-RRT})~\cite{kuffner_rrt-connect_2000}. As discussed earlier, standard state-based steering is only viable for fully or redundantly actuated systems as finding profiles $\qpos(t)$ for underactuated systems that satisfy the dynamics exactly is non-trivial. Due to this restriction, we instead adopt the control-based steering approach of \cite{kuffner_rrt-connect_2000} for our KNN-RRT implementation. Given UA1-RRT also uses a sample-based steering approach, we believe this choice enables a fair comparison.

All tests were run using a nearest neighbours value of $k=10$ in \textproc{Nearest}${}_k$ to avoid excessively long run times. For each method, we used the same random seed sequences for point selection in \textproc{RandomConfiguration}. We performed 20 seeds for the UAV example and 10 for the acrobot. We set a maximum run time of 5000 seconds for each run. For the path-parameterised methods, the integration procedures in $s$ used a resolution of $\Delta s = 10^{-3}$ and for KKN-RRT we used a time step $\Delta t = 10^{-2}s$. For each \textproc{Extend} procedure in UA1-RRT and KNN-RRT, we trialled $N_{rndm} = 200$ random paths/controls respectively, whereas in AVP-RRT we performed one extension similar to their original approach. We recorded the computation time for each \textproc{Steer} action within \textproc{Extend}, with average steering times $t_{\textproc{Steer}}$ recorded for each run of the examples.

\section{Results and Discussions}
All tests were performed on an Intel i7-2600 3.40 GHz processor, with the statistics for the UAV and acrobot tabulated in Tables \ref{tab: uav results} and \ref{tab: acrobot results} respectively. It is clear that of the three methods, UA1-RRT achieves the lowest mean run time and highest success rates in computing feasible trajectories for both examples.

\begin{table*}[t]
\begin{center}
\begin{tabular}{c|c|c|c|c|c|c}
\hline
 & Success (\%) & Run Time (s) & Feasible Edges (\%) & $t_{\textproc{Steer}}$ (ms) & Iterations & Vertices \\
\hline
UA1-RRT & \textbf{100.0} & \textbf{229.9 / 1214.0} & \textbf{100.0 / 100.0} & 1.00 / 1.29 & 11619 / 38571 & 10133 / 31619\\
AVP-RRT &  0.0 & 3188.1 / 5000.0 & 10.5 / 16.8 & 1.1 / 1.18 & 72747 / 133692 & 22443 / 46262\\
KNN-RRT & 15.0 & 4474.2 / 5000.0 & \textbf{100.0 / 100.0} & \textbf{0.40 / 0.53} & 102283 / 121742 & 122739 / 146090\\
\hline
\end{tabular}
\caption{UAV fly-through results and statistics (mean / max).}
\label{tab: uav results}
\end{center}
\end{table*}
\begin{table*}[t]
\begin{center}
\begin{tabular}{c|c|c|c|c|c|c}
\hline
 & Success (\%) & Run Time (s) & Feasible Edges (\%) & $t_{\textproc{Steer}}$ (ms) & Iterations & Vertices \\
\hline
UA1-RRT & \textbf{100.0} & \textbf{686.7 / 1886.7} & \textbf{100.0 / 100.0} & \textbf{0.10 / 0.15} & 37282 / 69581 & 26933 / 50329\\
AVP-RRT & 0.0 & 5000.0 / 5000.0 & 34.5 / 54.8 & 0.8 / 0.82 & 103991 / 291111 & 795 / 2305\\
KNN-RRT & 60.0 & 2076.5 / 5000.0 & \textbf{100.0 / 100.0} & 2.0 / 2.3 & 29440 / 69351 & 52990 / 124831\\
\hline
\end{tabular}
\caption{Acrobot swing-up results and statistics (mean / max).}
\label{tab: acrobot results}
\end{center}
\end{table*}

% Difficulty of AVP-RRT
Evaluation of AVP-RRT's resulting trajectories using \textproc{ComputeProfile} (Alg. \ref{alg: steer} L\ref{alg:line steer path profile}) were found to be only partially feasible, with trees containing at most 54.8\% feasible branches across all runs (Table \ref{tab: acrobot results}). Infeasible branches in these trees often admitted much higher values for $\theta$ in subsequent branches, leading to greater difficulty in reaching states of rest, typically reaching the maximum run time with very large iteration counts. Compared with the 100\% branch feasibility from UA1-RRT and KNN-RRT, it appears that AVP-RRT in its current form is not well suited for UA1 systems despite accommodating such constraints~\cite{pham_admissible_2017}.

\begin{figure}[t]
   \centering
	\begin{subfigure}[b]{1.0\linewidth}
         \centering
         \includegraphics[width=1.0\linewidth]{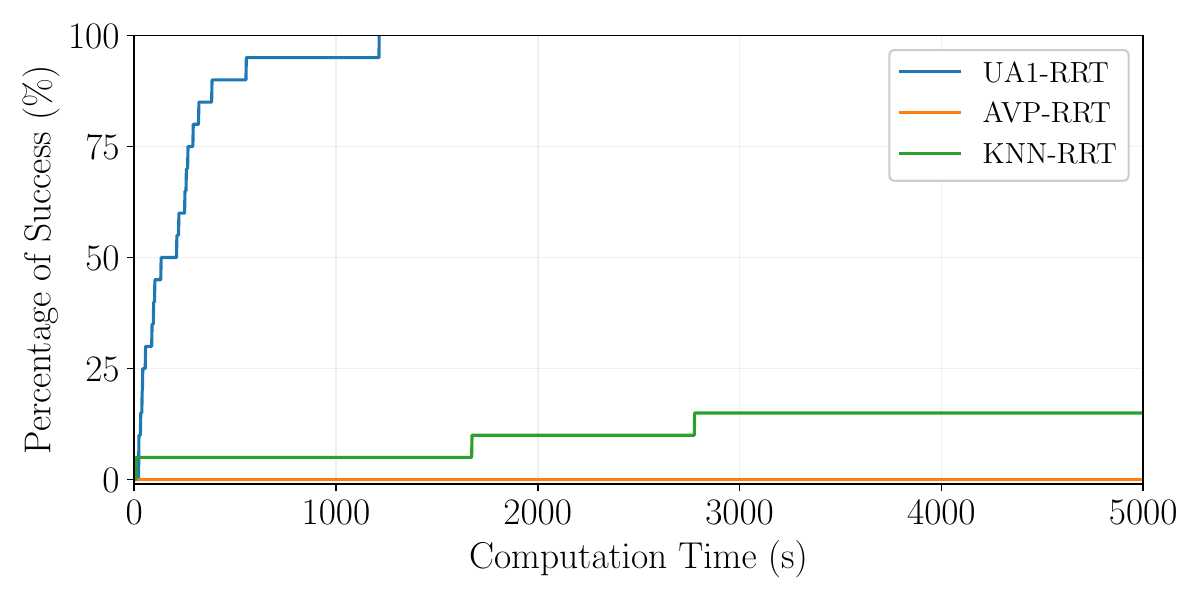}
         \caption{UAV fly-through.}
         \label{fig: percentage of success uav}
     \end{subfigure}\\   
     \begin{subfigure}[b]{1.0\linewidth}
         \centering
         \includegraphics[width=1.0\linewidth]{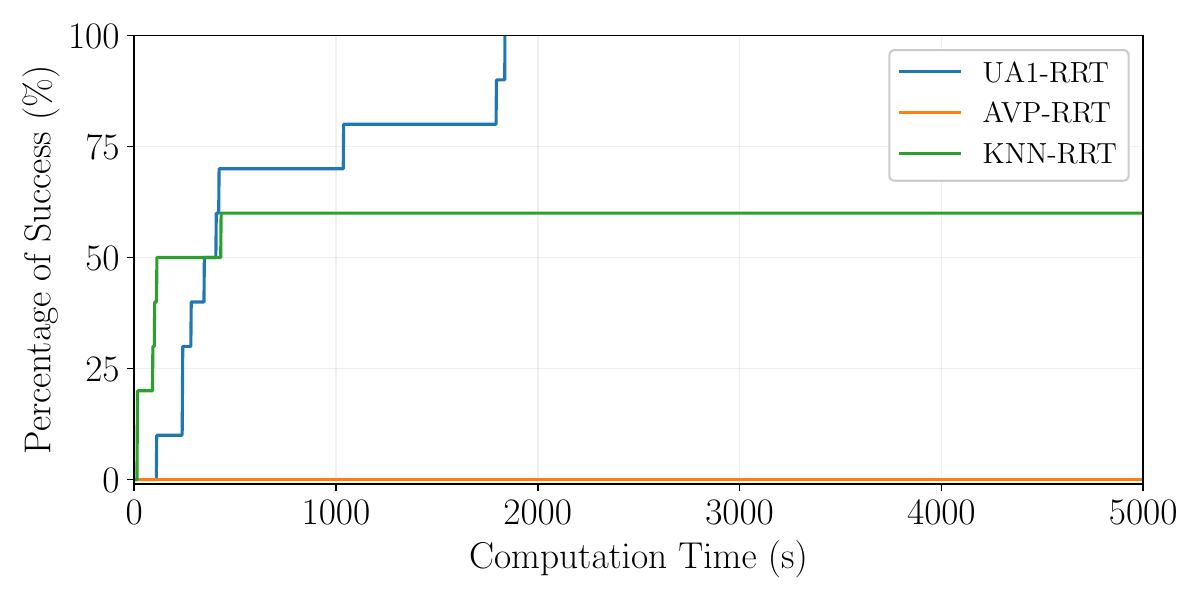}
         \caption{Acrobot swing-up.}
         \label{fig: percentage of success acrobot}
     \end{subfigure} 
    \caption{Percentage of successful attempts vs. computation time for the UAV and acrobot examples.}
    \label{fig: percentage of success}
\end{figure}

\begin{figure}[t]
    \centering
    \includegraphics[width=1.0\linewidth]{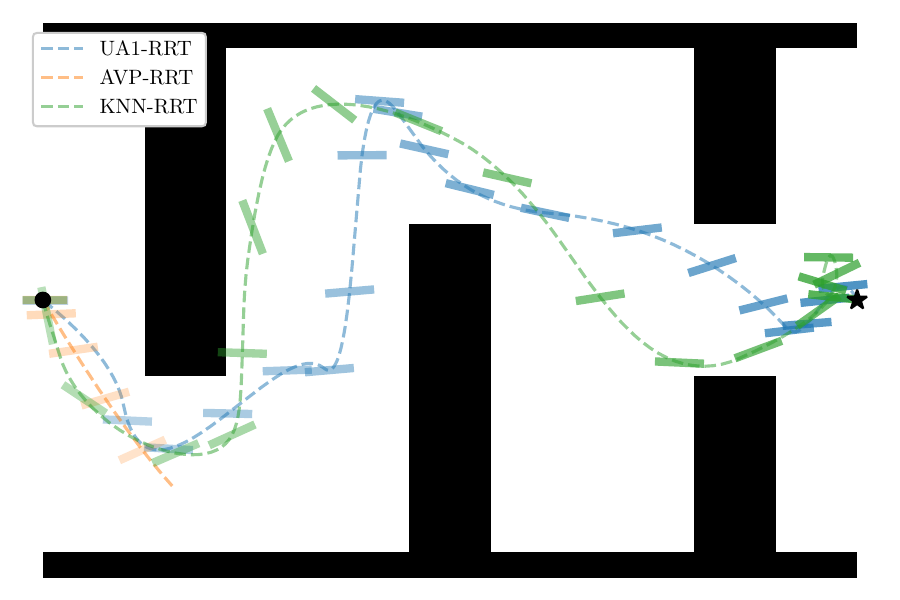}
    \caption{UAV trajectories found by each method \correctionadj{(feasible components of resulting trajectories shown)}.}
    \label{fig: uav motion}
\end{figure}

% Smoothness and its effect on tree growth
Fig. \ref{fig: percentage of success} shows the percentage of successful attempts achieved in each example over time, with UA1-RRT achieving greater success than KNN-RRT in both Fig. \ref{fig: percentage of success uav} and \ref{fig: percentage of success acrobot}. This could be attributed to the path smoothness encouraged by UA1-RRT's steering, enabling the growth of steady and feasible motions towards the goal whereas KNN-RRT's more aggressive approach will often create branches from which motion to the goal can not be recovered, leading to the higher failure rate and greater computational time. Fig. \ref{fig: uav motion} confirms this for the UAV, with the path-parameterised approaches generating visually smoother motions for the UAV in comparison to the sharper movements from the random-control steering of KNN-RRT, particularly in pitch angle. 

\begin{figure}[t]
    \centering
    \begin{subfigure}{0.23\textwidth}
         \centering
         \includegraphics[width=0.87\linewidth]{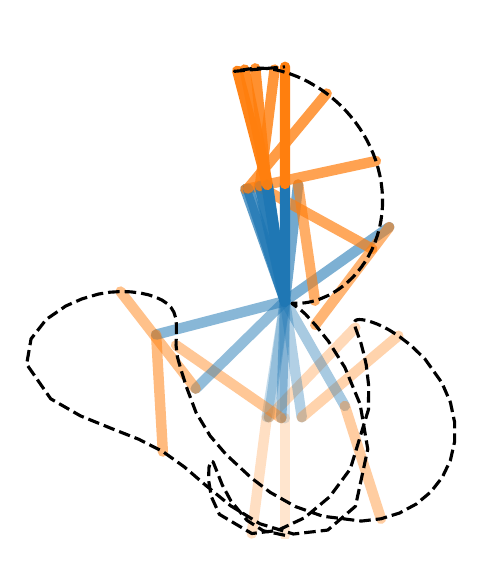}
         \label{fig: acrobot motion normal}
     \end{subfigure}    
     \begin{subfigure}{0.23\textwidth}
         \centering
         \includegraphics[width=0.65\linewidth]{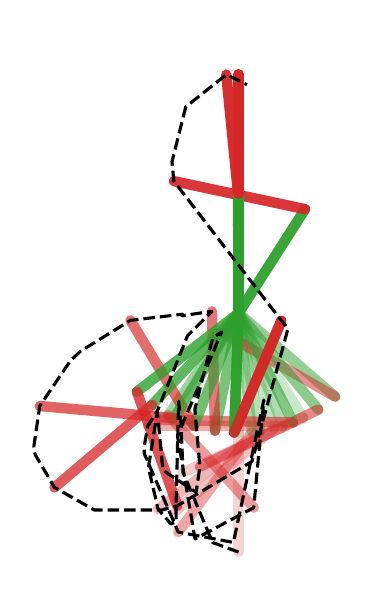}
         \label{fig: acrobot rrt motion}
     \end{subfigure}\\
    \begin{subfigure}{0.23\textwidth}
         \centering
         \includegraphics[width=1.0\linewidth]{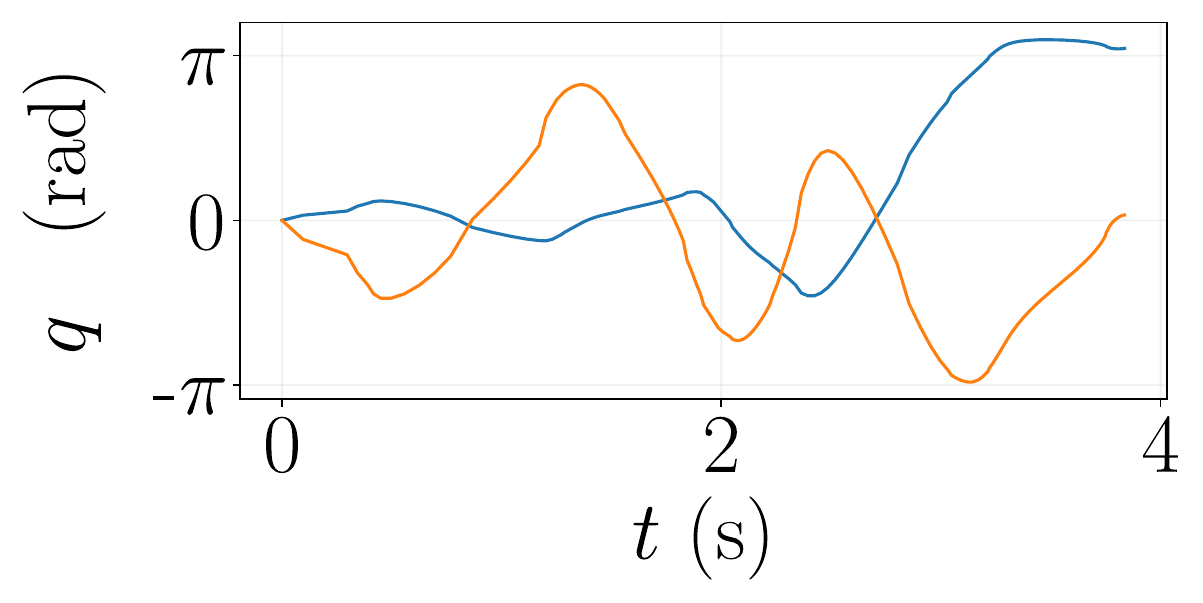}
         % \caption{UA1-RRT $\qpos(t)$}
         \label{fig: acrobot q}
     \end{subfigure}    
     \begin{subfigure}{0.23\textwidth}
         \centering
         \includegraphics[width=1.0\linewidth]{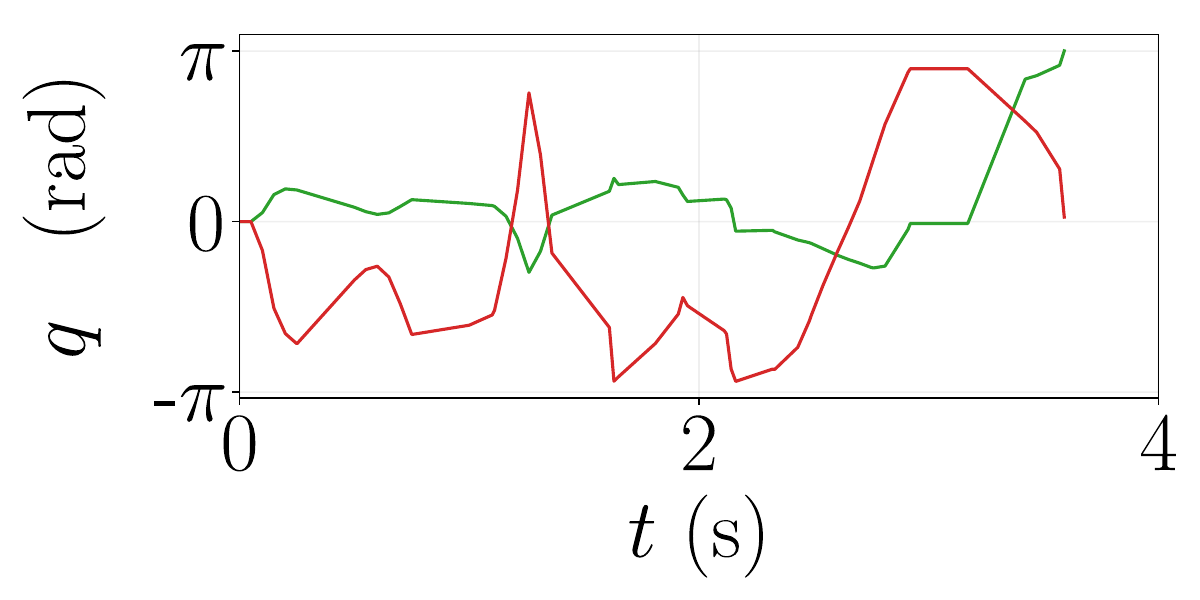}
         % \caption{UA1-RRT $\qpos(t)$}
         \label{fig: acrobot rrt q}
     \end{subfigure}\\
     \begin{subfigure}{0.23\textwidth}
         \centering
         \includegraphics[width=1.0\linewidth]{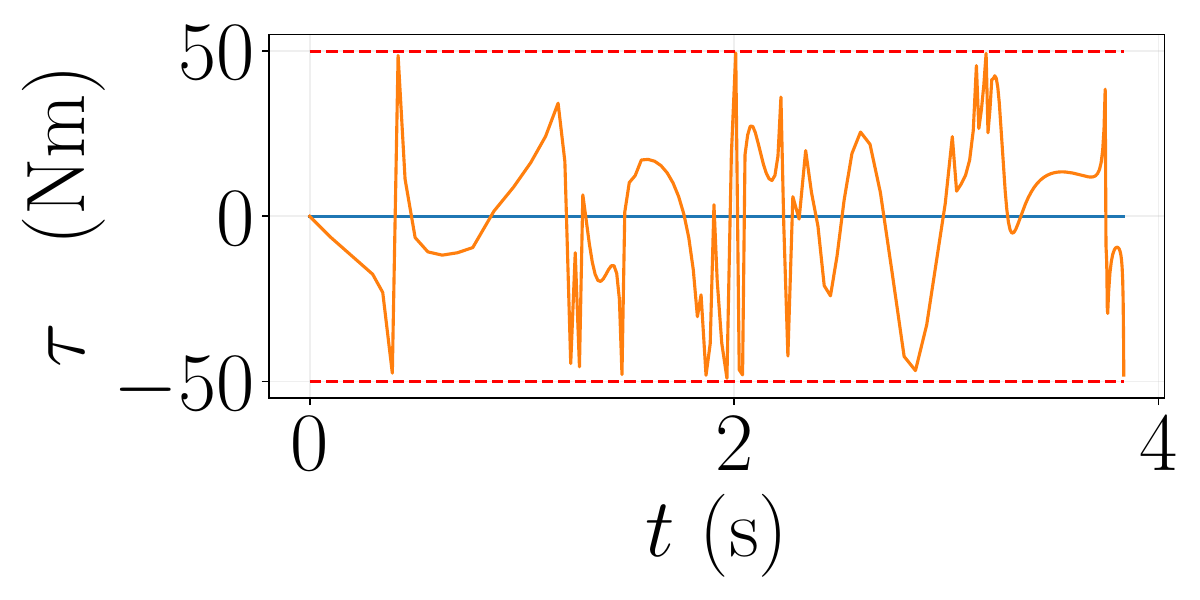}
         % \caption{KNN-RRT $\qpos(t)$}
         \label{fig: acrobot u}
     \end{subfigure} 
     	\begin{subfigure}{0.23\textwidth}
         \centering
         \includegraphics[width=1.0\linewidth]{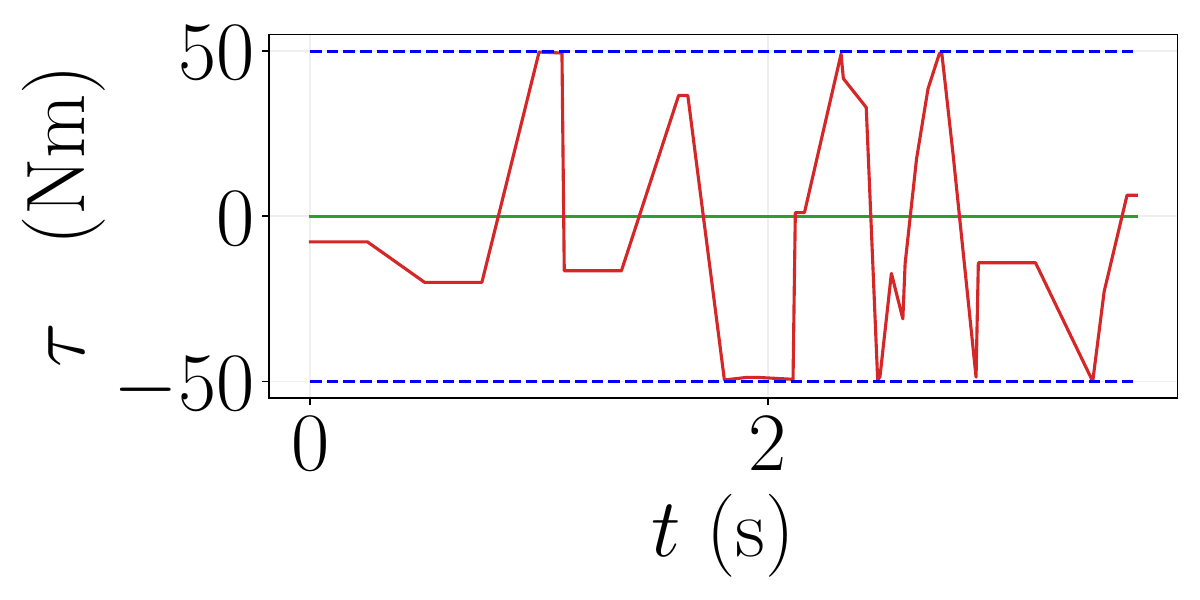}
         % \caption{KNN-RRT $\qvel(t)$}
         \label{fig: acrobot rrt u}
     \end{subfigure} 
     \caption{Example acrobot motions for UA1-RRT (left) and KNN-RRT (right) with profiles $\qpos(t)$ and $\gctrl(t)$.}
    \label{fig: acrobot motion}
\end{figure}

In the acrobot case (Fig. \ref{fig: percentage of success acrobot}), it is clear that KNN-RRT rapidly achieves 60\% of program success before stalling, whereas UA1-RRT's rate of success grows more steadily and reaches 100\% success. This difference also offers the insight that path smoothness could also impede tree growth if aggressive maneuvres are required of a system to continue tree expansion. Fig. \ref{fig: acrobot motion} illustrates an example in which our smooth path construction leads to conservative motion, where we observe the requirement of several build-up swings before the final swing-up is achieved. Comparing this with KNN-RRT, more aggressive motions are generated by the random-control steering, where the upright state is achieved in shorter times through stronger torques being held for longer durations, as shown in Fig. \ref{fig: acrobot motion}. The ability of KNN-RRT to move much quicker through areas of its state space (e.g. the final second of motion in Fig. \ref{fig: acrobot motion}) may explain the shorter completion times to UA1-RRT in \ref{fig: percentage of success acrobot}, where the expansion of the tree through more varied motions can reach the goal in shorter times. To increase our approach's flexibility towards such erratic behaviour, $D_{\max}$ can be decreased appropriately, at the expense of denser and ultimately slower tree growth. 

% Steering times
The rapid detection of infeasibility in \textproc{PathProfile} allows UA1-RRT to discard paths with $s^\star < s^\dagger$ immediately in \textproc{Steer} (Alg. \ref{alg: steer}). As a result, little computation time is spent on paths created by \textproc{GeneratePath} that are not feasible. This is beneficial for systems with strong couplings between their paths and dynamics, as the majority of generated paths will be entirely infeasible and will not contribute to computational overhead. This is particularly the case for the acrobot, with steering times for UA1-RRT being the lowest by a considerable margin (Table \ref{tab: acrobot results}). AVP-RRT on the other hand performs several orders of magnitude slower for both examples, when considering UA1-RRT and KNN-RRT perform up to 200 trajectory evaluations per call of \textproc{Steer} whereas AVP-RRT only performs one. This is expected, however, since AVP must pre-compute several components such as limiting curves and switch points~\cite{pham_general_2014} to generate their velocity profiles, which UA1-RRT and KNN-RRT need not consider and instead compute theirs with a single forward pass.

\section{Conclusions}
We have presented a sample-based motion planning algorithm using path parameterisation specialised to underactuated degree-one systems. Using the structure these systems present under this parameterisation, an efficient state-based steering method was developed. For the examples considered in this paper, our proposed algorithm UA1-RRT demonstrated much higher rates of success and shorter mean run times in computing feasible trajectories in comparison to KNN-RRT and AVP-RRT, where these existing methods found difficulty. We highlighted that these achievements were largely attributed to our state-based steering approach, with efficient computations achieved without introducing excessive computational overhead. 

Potential future avenues of research  include an investigation into the scalability of UA1-RRT with models of higher dimension \correctionnew{as well as those operating in more complex, obstacle-filled environments}.

\addtolength{\textheight}{-12cm}   % This command serves to balance the column lengths
                                  % on the last page of the document manually. It shortens
                                  % the textheight of the last page by a suitable amount.
                                  % This command does not take effect until the next page
                                  % so it should come on the page before the last. Make
                                  % sure that you do not shorten the textheight too much.

\bibliographystyle{./bibliography/IEEEtran}
\bibliography{./bibliography/IEEEabrv,./bibliography/references}

\end{document}